\pdfoutput=1
\documentclass[11pt]{article}

\usepackage[]{acl}

\usepackage{times}
\usepackage{latexsym}
\usepackage{comment}

\usepackage[T1]{fontenc}
\usepackage[utf8]{inputenc}
\usepackage{listings}
\usepackage{multicol}
\usepackage{multirow}
\usepackage{graphicx}
\usepackage{booktabs}
\usepackage{float}
\usepackage{soul}
\usepackage{amsmath}
\usepackage{enumitem}
\usepackage{times}
\usepackage{latexsym}
\usepackage{graphics}
\usepackage{multirow}
\usepackage{inconsolata}
\usepackage{todonotes}

\usepackage{listings} % Add this for using the listings package
\usepackage{graphicx}
\usepackage{subcaption}
\usepackage{verbatim}
\usepackage{setspace}

\usepackage{arabtex}
\usepackage{utf8}
\setcode{utf8}

\newcommand{\ds }{\emph{ThatiAR}} %SubjecTAR

\title{\ds{}: Subjectivity Detection in Arabic News Sentences}

\author{Reem Suwaileh$^1$$^*$, Maram Hasanain$^2$, Fatema Hubail$^3$, Wajdi Zaghouani$^1$, Firoj Alam$^2$\thanks{~~Equal contribution.}\\  
  $^1$Hamad bin Khalifa University, Qatar\\
  $^2$Qatar Computing Research Institute, HBKU, Qatar\\
  $^3$Free University of Berlin, Germany\\
  {\tt \{rsuwaileh,mhasanain,wzaghouani,fialam\}@hbku.edu.qa}, fatemahubail93@gmail.com \\
\\}

\begin{document}
\maketitle

\begin{abstract}
Detecting subjectivity in news sentences is crucial for identifying media bias, enhancing credibility, and combating misinformation by flagging opinion-based content. It provides insights into public sentiment, empowers readers to make informed decisions, and encourages critical thinking. While research has developed methods and systems for this purpose, most efforts have focused on English and other high-resourced languages.
In this study, we present the first large dataset for subjectivity detection in Arabic, consisting of $\sim$3.6K manually annotated sentences, and GPT-4o based explanation. In addition, we included instructions (both in English and Arabic) to facilitate LLM based fine-tuning. We provide an in-depth analysis of the dataset, annotation process, and extensive benchmark results, including PLMs and LLMs. Our analysis of the annotation process highlights that annotators were strongly influenced by their political, cultural, and religious backgrounds, especially at the beginning of the annotation process. The experimental results suggest that LLMs with in-context learning provide better performance. We aim to release the dataset and resources for the community.\footnote{\url{anonymous.com}} 
% Subjectivity and Sentiment Analysis (SSA) has gained significant attention among researchers due to its wide applications. While the majority of studies has focused on English, the field is expanding to incorporate complex languages like Arabic, and Turkish. The first step in developing systems for SSA is usually creating an annotated dataset for the task. During the course of developing such dataset for subjectivity prediction, we observed that the annotators personal characteristics affected their judgments. Thus, in this work, we aim to understand whether annotators can \textit{objectively} label for the task within the context of Arabic news reporting. Our research further asks if current annotation guidelines should be improved to account for the morphologically-rich, socially complex, and culturally-specific content. We answer these questions through an extensive manual and statistical investigation of an in-house subjectivity annotation effort of sentences from Arabic news articles. The results highlight that annotators for the task were strongly-influenced by their political, cultural and religious backgrounds especially in the beginning of the annotation process. 
\end{list}
\end{abstract}

\section{Introduction}
\label{sec:introduction}

\begin{figure}[]
    \centering
    \includegraphics[scale=0.27]{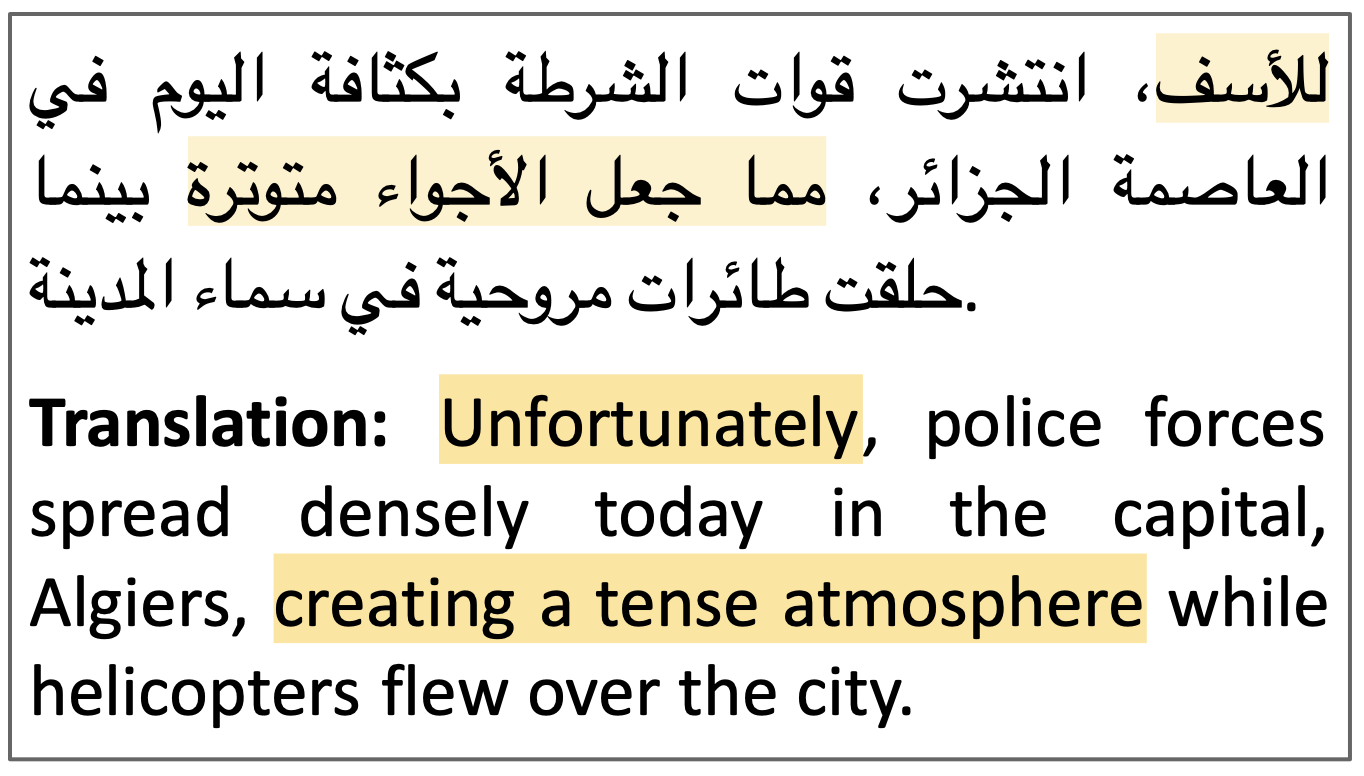}
    \caption{An example of a subjective sentence that can be misleading and cause fear.}
    \label{fig:arabic_subj_example}
    \vspace*{-0.2cm}
\end{figure}

Detecting subjectivity\footnote{Subjectivity, according to \citet{abo2019ssa}, `` refers to aspects of language used to express feelings, opinions, evaluations, and speculations and, as such, it incorporates sentiment''.} in news sentences is crucial for several reasons. It helps identifying media bias by distinguishing between objective reporting and subjective content, thereby enhancing the credibility of news sources. This differentiation is also vital in combating misinformation and fake news by flagging opinion-based content for further verification. In Figure \ref{fig:arabic_subj_example}, we present an example of a subjective sentence that can be misleading and cause fear among citizens. The highlighted part of the text in the example is subjective. 

With the reliance on social media as platforms of expression, users often resort to informality, dialects, and a combination of languages. When seeking news reports and statements, readers turn to news outlets for knowledge and assessments of current events. While readers may consider news from reliable outlets as objective sources of information, research shows that news reports are often partisan, subjective, and reflective of the news agency's standpoint. Therefore, analyzing subjectivity provides insights into public sentiment and the social impact of news. It can empower readers to make informed decisions and encourages critical thinking by highlighting subjective content in news articles.

While there has been research effort to develop methods and systems to automatically identify such content, the majority of studies focus on English or other high-resourced languages. However, the field is growing to incorporate ``morphologically-rich'' or complex languages, including Urdu, Arabic, and Turkish~\cite{abdul2011acl, abdul2011subjectivity, abdul2015subjectivity}.

% most of such efforts have focused on English and other high-resourced languages.
% With the proliferation of data across the web and social media platforms, researchers have become significantly interested in mining this massive data to further assess how end-users receive, understand, and share them. Central to this field is Subjectivity and Sentiment Analysis (SSA). 
% Subjectivity is a lexical resource that allows researchers to categorize polarity and sentiment embedded through texts \cite{al2019comprehensive}. While subjectivity incorporates sentiments, objectivity is usually considered as statements of facts, thus, devoid of sentiments. 
% The area of SSA
%Subjectivity and Sentiment Analysis (SSA) 
% is growing in academic literature pertaining to Computational Social Sciences, Digital Humanities, among other fields. With a 
% Subjectivity analysis allows researchers to annotate and categorize texts as objective or subjective, while sentiment analysis becomes a complementary approach providing a secondary analysis of subjective text often categorized as positive, negative, neutral, and mixed~\cite{al2014subjectivity}. 

Research on subjectivity in Arabic content addresses the complexities of language usage, primarily focusing on Modern Standard Arabic (MSA) and regional dialects. The significant variation in Arabic dialects across different geographical and national contexts presents an additional challenge for annotators from diverse backgrounds. Therefore, in this study, we focused on Arabic, with a special emphasis on news content. Given the lack of resources for developing AI-based systems in Arabic, we introduce the \ds,\footnote{Translated in Arabic as \<\footnotesize ذاتي> (``Thati'') meaning ``subjective'' in English.} a large and reasonably well balanced dataset consisting of manually annotated news sentences. While doing manual annotation, we also focused on following research questions: \textit{(i)} What are the emerging characteristics of news reports with regards to subjectivity? \textit{(ii)} How do annotators of diverse backgrounds approach news reports? and, \textit{(iii)} Should current subjectivity annotation strategy and guidelines be further developed to account for morphologically-rich, socially complex, and culturally-specific content? 
We conducted extensive experiments to create a benchmark using different Pre-trained Language Models (PLMs) and Large Language Models (LLMs) that can serve as a foundation for future research. Given that current LLMs consistently push the boundaries of NLP and achieve state-of-the-art performance in tasks such as machine translation, summarization, sentiment analysis, and more complex applications like legal document analysis and creative writing \citep{liang2022holistic,openai2023gpt4,bang2023multitask,ahuja2023mega,hendy2023good,tawkat2023gptaraeval}, therefore, we used GPT-4o to generate explanations for why a sentence is labeled as subjective or objective. Additionally, we developed instructions for each data point, resulting in a comprehensive instruction-following dataset.
Below we provide a summary of our contributions, 
\begin{itemize}[noitemsep,topsep=0pt,leftmargin=*,labelwidth=!,labelsep=.5em]
    \item We developed \ds, a dataset consisting of approximately 3.6K manually annotated news sentences. This is largest dataset compared to any other subjectivity dataset released so far. 
    \item We provide a detailed analysis of the annotation process, addressing the research questions mentioned earlier.
    \item Benchmark results using different PLMs and LLMs.
    \item The dataset includes explanations for the provided labels, which can aid in developing explanation-based generative models.
    \item An instruction-following dataset that can help in building models capable of following instructions.
\end{itemize}
    
% \rs{add paper structure}
The remainder of the paper is structured as follows: Section \ref{sec:related_work} provides an overview of relevant literature. Section \ref{sec:dataset} details the dataset used, along with a detailed analysis. In Section \ref{sec:experiments}, we discuss the models and experiments. The results and findings are presented in Section \ref{sec:results}. We developed explanation and instruction datasets, which are discussed in Sections \ref{sec:data_rationale} and \ref{sec:data_instruct}. Finally, in Section \ref{sec:conclusion} we provides concluding remarks.

\section{Related Work}
\label{sec:related_work}

% Subjectivity analysis allows the classifying of texts as subjective or objective.  In conjunction with subjectivity analysis, sentiment analysis, according to~\citet{al2014subjectivity}, is used ``as the second level of classifying subjective text into positive or negative.''  Researchers often approach subjectivity and sentiment analysis hierarchically where texts are classified as subjective or objective, and then designating sentiments as ``positive, negative, or mixed" to subjective texts~\cite{korayem2012subjectivity,mourad2013subjectivity,refaee2014a,refaee2014b}. Alternatively, researchers approach the text directly and begin classifying it on the basis of sentiment.

Research on subjectivity analysis often approaches subjectivity and sentiment analysis hierarchically. First, texts are classified as subjective or objective, and then sentiments are designated as ``positive,'' ``negative,'' or ``mixed'' for the subjective texts~\cite{korayem2012subjectivity,mourad2013subjectivity,refaee2014a,refaee2014b}. Typically, it has been served as a preliminary step to sentiment analysis \cite{savinova-moscoso-del-prado-2023-analyzing}, as it relies primarily on subjective fragments of the text. Earlier approaches of research for this domain was mainly rule based and mostly for English. Recently the problem has been mostly addressed by training transformer based models \cite{huo2020utilizing}. 

% With regards to the Arabic language, the field of subjectivity analysis has widened as researchers continue to develop corpora including both MSA and Dialectal Arabic (DA). 
For Arabic, earlier research by \citet{abdul2014samar} proposed a system for sentence-level subjectivity analysis of Arabic social media. They also developed a comprehensive corpus that includes sentences from chats, tweets, Wikipedia pages, and web forums, which were manually annotated as objective, subjective, neutral, or mixed, and further categorized by sentiment (i.e., positive and negative). \citet{habash2013qalb} developed the Qatar Arabic Language Bank (QALB), which provides guidelines for Arabic corpus annotations that account for the Qatari dialect. These corpora and dataset developments are significant to the field of subjectivity and sentiment analysis in both Modern Standard Arabic (MSA) and Dialectal Arabic (DA). Additionally, \citet{azmi2014aara} developed an opinion mining system targeting the Saudi Najdi Dialect, called Ara’a. This dataset includes comments manually annotated for sentiment polarities. 

The development of AI-based systems requires annotated datasets. The dataset development with subjectivity annotations are inherently subjective and influenced by annotators' standpoints, social contexts, backgrounds, and other factors. Additionally, political stances can affect how annotators interpret and annotate the text~\cite{luo2020detecting,diaz2018addressing}. This introduces a significant gap in the emerging literature on subjectivity, particularly within the diverse Arabic linguistic context.

The implications of manual annotations in subjectivity detection are challenging, reflecting the inherently subjective nature of the task. Forms of agreement and disagreement among annotators provide insights into the subjective nature of the content and highlight the challenges in achieving consistent annotations. High agreement levels indicate clearer subjective or objective content, while disagreements reveal areas where subjectivity is more ambiguous and contested.

Such findings can show a gap in the literature on the development of subjectivity detection systems for the Arabic language. Addressing this gap requires a deeper understanding of how these factors impact annotations and the development of more robust, context-aware AI systems.

Considering different challenges and aspects, in this study we propose \ds{}~dataset and provides benchmark results, which is a first of its kind for Arabic. This resource will benefit the community towards building models focusing on small and large models and conduct further resarch for news media analysis.

\begin{figure*}[ht!]
    \centering
    \includegraphics[scale=0.35]{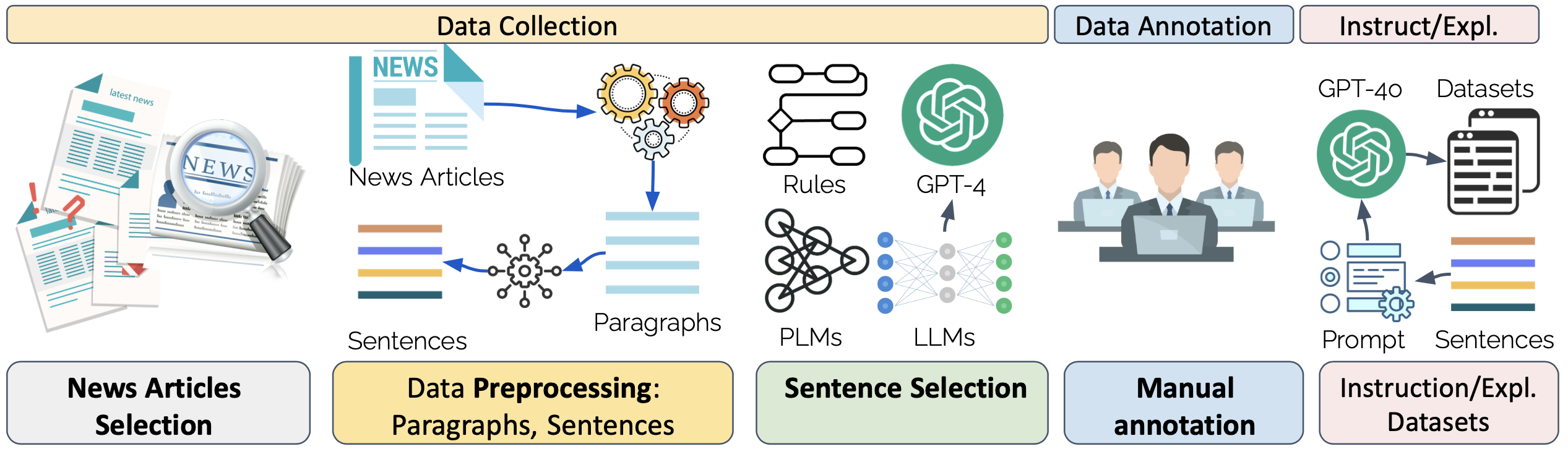}    
    
    \caption{The pipeline of the data collection, annotation, and instruction/explanation datasets development process.}
    \label{fig:data_collection_pipeline}
    \vspace{-0.3cm}
\end{figure*}

\section{Dataset} 
\label{sec:dataset}
In this section, we discuss the effort carried out to construct \ds~dataset. 
% including the data collection (Section ~\ref{sec:data_collect}), data annotation (Section ~\ref{sec:data_annotate}), and data analysis (Section ~\ref{sec:data_analyze}). 
In Figure \ref{fig:data_collection_pipeline}, we provide a complete workflow of the data collection (Section ~\ref{sec:data_collect}), manual annotation (Section ~\ref{sec:data_annotate}) and instruction/explanation (see Sections \ref{sec:data_rationale} and \ref{sec:data_instruct}) procedures.

%%%%%%%%%%%%%%%%%%%%%%%%%%%%%%%%%%%%%%%%%%
\subsection{Data Collection} 
\label{sec:data_collect}
To prepare a set of sentences for subjectivity annotation, we went through the following two phases.

\subsubsection{News Article Selection} 
We selected the AraFacts dataset \cite{sheikh-ali-etal-2021-arafacts}, which contains claims verified by Arabic fact-checking websites. Each claim is associated with web pages that either propagate or refute the claim. In total, we collected 1,159 new articles from AraFacts. To address the issue of skewed distribution of sentence types in news articles, which tend to favor objective sentences, we enlisted a graduate student to manually search for opinionated articles published by various Arabic news outlets (e.g., Sky News Arabia, Alarabiya, Al-Masry Al-Youm). This effort resulted in selecting 221 new articles.

\subsubsection{Preprocessing} 
We parsed the web pages using three different scrapers, favoring the longest output. The tools used were Goose3,\footnote{\url{https://goose3.readthedocs.io}} Newspaper3k,\footnote{\url{https://newspaper.readthedocs.io/}} and Trafilatura.\footnote{\url{https://trafilatura.readthedocs.io}} After extracting the text content, we segmented the body of text into paragraphs and sentences, resulting in a total of 15,947 sentences. The parsing and segmentation involves rule based approaches to filter and remove noisy html tags.

%%%%%%%%%%%%%%%%%%%%%%%%%%%%%%%%%%%%%%%%%%
\subsubsection{Sentence Selection} 
We applied two sampling strategies. The first strategy was rule-based, considering only sentences with a length between 10 and 45 words to select not overly long self-contained sentences. The second strategy utilized four transformer-based models and GPT-4 to evaluate the subjectivity of the sentences. The goal of this strategy was to select sentences with at least one ``subjective'' vote, thereby oversampling potentially subjective sentences for annotation. 
For this purpose, we trained five models: ARABERTv2 \cite{antoun2020arabert}, ARBERTv2 \cite{abdul-mageed-etal-2021-arbert}, MARBERTv2 \cite{abdul-mageed-etal-2021-arbert}, GIGABERTv4 \cite{lan2020gigabert}, and GPT-4 \cite{achiam2023gpt}. These models were fine-tuned on the entire Arabic subjectivity datasets from the CheckThat! 2023 lab \cite{clef-checkthat:2023:task2}. As a result, we selected 4,524 sentences, forming our annotation pool. 

%%%%%%%%%%%%%%%%%%%%%%%%%%%%%%%%%%%%%%%%%%
\subsection{Data Annotation} 
\label{sec:data_annotate}
To annotate \ds, we opt to employ human annotators on Amazon Mechanical Turk (mTurk) crowdsourcing platform. We adopted the annotation guidelines from a previous study \cite{Francesco21} and tailored them for Arabic language. We discuss the annotation guidelines with additional examples in Appendix~\ref{apndix:guidelines}. 
%\firoj{We can briefly discuss the annotation instructions here, with a subsecion. Also we can add Arabic instruction in appendix, might be useful for the community. Do you have the word file?} \rs{will (1) update the Appendix section with our guidelines and (2) add here brief instructions if we have space ..}
In a nutshell, we define \emph{subjective} sentences as expressions of the writer's feelings, literary tastes, or personal interpretations of topics and events. Sentences containing sarcasm, humor, support, or offensive language are also considered subjective. In contrast, \emph{objective} sentences present facts, events, and topics based on verifiable data and include common expressions or sayings not originally written by the author.

To ensure the clarity and coherence of the guidelines, and the mTurk annotation configuration, we ran multiple pilot studies that exhausted around 850 sentences from our pool. 

To ensure the quality of annotations, we prepared/sampled a set of 115 in-house annotated sentences from CheckThat! 2023 lab \cite{clef-checkthat:2023:task2}. We use these sentences in two ways: \textit{(i)} 10 questions for pre-qualification test that an annotator has to pass before being eligible to start the actual HITs, and \textit{(ii)} 105 questions for ongoing-qualification that an annotators has to maintain an acceptable accuracy throughout the annotation process. For both we requested the worker accuracy above 60\%.

We finally set up the design of the annotation interface and configurations as follows. We ran 245 HITs, each containing no more than 15 sentences and 5 quiz questions. We initially collected 3 annotations per sentence and dynamically requested up to 2 more annotations when the majority agreement of 66.6\% was not met, to guarantee the reliability of annotations. As a result, we obtained 3,661 sentences with 66.6\% agreement, of which 1,579 were subjective and 2,082 were objective sentences. The sentences that did not pass the agreement score were removed from the final dataset. In Table \ref{tab:examples}, we present a few annotated sentences from the \ds{} dataset along with their English translations. Table~\ref{tab:data_stats} shows the statistics of \ds{}.  
% We also show example sentences from \ds{} dataset in Table~\ref{tab:examples}

\begin{table}[]
    \centering
    \scalebox{0.85}{%
    \begin{tabular}{l rrr} \toprule
    \textbf{Set} & \textbf{SUBJ} & \textbf{OBJ} & \textbf{All} \\\midrule
    Train & 1,055 (66.8\%)  & 1,391	(66.8\%) & 2,446 \\
    Dev & 201 (12.7\%) & 266 (12.8\%) & 467 \\
    Test & 323	(20.5\%) & 425 (20.4\%) & 748 \\\midrule
    \textbf{All} & \textbf{1,579} & \textbf{2,082} & \textbf{3,661} \\\bottomrule
    \end{tabular}
    }
    \caption{Statistics of \ds{} dataset}
    \label{tab:data_stats}
    \vspace{-0.3cm}
\end{table}

\begin{table*}[]
    \centering
    \scalebox{0.9}{%    
    \begin{tabular}{ll p{6.4cm}p{6.6cm}} \toprule
    \#& Label & Sentence & Translation \\\midrule

    1 & \textsc{subj} 
        & \begin{RLtext} \footnotesize وجدت بوحيرد نفسها -وهي فتاة تبلغ من العمر 22 عاما - بين يدي ضباط المستعمر الفرنسي ينهش لحمها بكل الطرق. \end{RLtext} 
        & Bouhired found herself - a 22-year-old girl - in the hands of French colonial officers, a prey whose flesh was being devoured in every way. \\[.2cm]
        
    2 & \textsc{subj}  
        & \begin{RLtext} \footnotesize ولكنني لم أجد الوقت الكافي للتعرف عليه عن كثب ولكن عندما مررت بأوقات عصيبة، أعطتني العقيدة الإسلامية القوة اللازمة لمواجهته. \end{RLtext} 
        & But I did not find the time to get to know Islam closely, but when I went through difficult times, the Islamic faith gave me the strength necessary to face COVID. \\[.2cm]
    
    3 & \textsc{obj} 	  
        & \begin{RLtext} \footnotesize كما تدخل نترات الأمونيوم في صناعة المتفجرات خاصة في مجال التعدين والمناجم.\end{RLtext} 
        & Ammonium nitrate is also used in the manufacture of explosives, especially in the field of mining. \\[.2cm]
    
    4 & \textsc{obj}    
%        & \<\footnotesize السيناتور الجمهوري بات تومي يحمل ترمب المسؤولية>
%        \<\footnotesize  عن احتجاجات واشنطن ويصفه بـ `الغوغائي'.> 
%        & Republican Senator Pat Toomey holds Trump responsible for the Washington protests and describes him as a ``demagogue.'' \\
        & \begin{RLtext}\footnotesize اشتية: السعوديون أعادوا القضية الفلسطينية للطاولة عند الحديث عن التطبيع مع إسرائيل \end{RLtext}     
        & Shtayyeh: The Saudis put the Palestinian issue back on the table when talking about normalization with Israel \\
        \bottomrule

    \end{tabular}
    }
    \caption{Example sentences from \ds{} dataset. 
    % \rs{we can remove this table}
    }
    \label{tab:examples}
    \vspace{-0.5cm}
\end{table*}

%%%%%%%%%%%%%%%%%%%%%%%%%%%%%%%%%%%%%%%%%%
\subsection{Data Analysis} \label{sec:data_analyze}
%%%%%%%%%%%%%%%%%%%%%
% \paragraph{Reliability.}
\paragraph{Annotation Agreement.}
To evaluate the reliability of human annotations, we computed the Inter-Annotator Agreement (IAA) using an agreement coefficient
% \footnote{\url{https://www.nltk.org/_modules/nltk/metrics/agreement.html#AnnotationTask.avg_Ao}} 
that averages the observed agreement across all annotators and sentences. We found the agreement to be approximately 0.54, indicating a reasonable level of agreement for the subjectivity annotation.

We further computed the Cohen's Kappa (C.Kappa) coefficient between each of the first three annotators and the consolidated label (determined by majority voting)  \cite{alam-etal-2021-fighting-covid}. As shown in Table~\ref{tab:iaa}, the C.Kappa results indicate a moderate agreement, with an average of 0.54.\footnote{According to Landis and Koch's scale \cite{landis1977measurement}, Kappa values of 0.21–0.40 correspond to fair agreement, 0.41–0.60 to moderate agreement, 0.61–0.80 to substantial agreement, and 0.81–1.0 to perfect agreement.} The annotation task for subjectivity is complex, which effects the agreement score. This complexity has also been highlighted in \cite{Francesco21,antici-etal-2024-corpus-sentence}. 

\begin{table}[ht!]
    \centering
    \scalebox{0.9}{%
    \begin{tabular}{lr} \\\toprule
    \textbf{Setup} & \textbf{C.Kappa} \\\midrule
    Annotator1 vs. Majority & 0.5464 \\
    Annotator2 vs. Majority & 0.5512 \\
    Annotator3 vs. Majority & 0.5173 \\\midrule
    \textbf{Average} & 0.5383 \\\bottomrule
    \end{tabular}
    }
    \caption{Inter-annotator agreement using Cohen Kappa ($\alpha$) for \ds{} dataset}
    \label{tab:iaa}
\end{table}
%

%%%%%%%%%%%%%%%%%%%%%
%\paragraph{Diversity and Coverage}
%\ds{} contains a wide range of news outlets, covering more than 500 news outlets. Figure~\ref{fig:domain} depicts the top frequent domains in \ds{}. %AraFacts 490 domains and Opinionated 12 domains ..

%%%%%%%%%%%%%%%%%%%%%
%\paragraph{challenges}
%\firoj{We can bring Annotation challenges from appendix to here}
%\rs{better to keep them in Appendix after guidelines. Please note that the challenges are related to the English guidelines used last year. This year we refined them to Arabic language ...}

%\subsubsection{Annotation Deep Analysis}
\paragraph{Deep Analysis.}
While manually annotating \ds{}, we focused on the key aspects that impact the understanding of Arabic news reporting and the quality of annotations. We discuss our analysis by discussing the examples reported Table~\ref{tab:news_analysis} (in Appendix). 

\noindent \textit{\textbf{Bias in reporting and annotating.}} 
News reports often contain phrases and terms that can be interpreted in multiple ways. Sentence \#1 is example in point. The phrase ``\<\footnotesize الإقليم المضطرب>'' (``volatile region'') is a preliminary site of disagreement. The region may be described as volatile because it merits the description of Oxford dictionary definition: \textit{``liable to change rapidly and unpredictably, especially for the worse.''} However, the perception of volatility could also be influenced by partisan news reporting that portrays China as oppressive and democracy as liberating. This raises the question of whether the term ``volatile'' is accurate or if it carries political, historical, or cultural biases of the journalists and news agencies. On the other hand, annotators with similar potential biases are likely to consider this news sentence objective, while those with differing biases may view it as subjective.

%\noindent \textit{\textbf{Subjectivity annotation is subjective.}} 
\noindent \textit{\textbf{Subjectivity in reporting and annotating.}} 
To understand the sources of disagreement between annotators, we examined several instances that exhibits some aspects contributing to their subjectivity. For instance, sentence \#2 references ``\<\footnotesize الاحتلال>'' (``the occupation''), which readers commonly understand to mean ``Israel.'' This term is politically loaded and functions as a critique of the Israeli occupation, placing blame on Israel as an occupying power and alluding to other historical occupations. Additionally, the phrase \<\footnotesize اختراق الحدود> (``border breaching'') followed by ``\<\footnotesize الأراضي الفلسطينية>'' (``Palestinian lands'') also carries significant political weight. If this report were from a news agency that supports Israeli claims to nationhood, it might not use ``Palestinian lands'' or refer to Israel as ``the occupation.'' The term ``border breaching''  implies unlawful activity, indicating subjectivity in the portrayal of events. %Although the sentence might appear as a factual statement, it is inherently subjective. 
The subjectivity in this sentence may not intended as a negative or politically motivated claim but rather emerges from religious and cultural contexts that are more easily understood by regional annotators.

\noindent \textit{\textbf{Composite reporting.}} 
Multiple news sentences often report different matters within the same text segment. For example, sentence \#3 combines three distinct headlines into one statement, each containing both subjective and objective descriptions. This discrepancy can lead to disagreement among annotators, as each annotator may focus on different parts of the sentence or interpret the main focus differently.

\noindent \textit{\textbf{Perspectives of annotators.}} %Neutral and ambiguous reporting.}} 
%Not all sentences must be subjective or objective. Therefore, asking annotators to adhere strictly to the annotation guidelines, meaning matching texts to each case as presented in the annotation guidelines to further label the text as either subjective or objective, is a major attribution to the increase the level of disagreements. 
The perspective of annotators and their standpoint is a crucial element in their judgments. For instance, feminist annotator would account for terms in sentence \#4 ``\<\footnotesize قام بالتآمر>'' (``conspired'') and ``\<\footnotesize وهو عالم>'' (``knowing [well]''), and the usage of the term ``\<\footnotesize فإن>'' (``if'') and ``\<\footnotesize فسيتخلص>''  (``get rid of''), therefore judge it to stand collectively as a subjective sentence. The term ``conspire'' implies criminal or unlawful activities; knowing, a subjective term, espouses knowledge as more prevalent in one person than another; and ``get rid'' implies the parents, although violent, are disposable further dehumanizing them through the lens of criminal activity. Had this statement appeared in English, the statement would appear immediately subjective. However, in the construction of the sentence in Arabic, the initial clause functions as a factual statement, further justified by the following clause, and that subjectivity may only be interpreted as appearing in the last line with the term ``rid''.

To this end, we answer our three questions in light of our examination and analysis of several cases and instances:
\begin{enumerate}[itemsep=0pt, topsep=0pt, partopsep=0pt, parsep=0pt]
    \item[Q1.] \textit{What are the emerging characteristics of news reports with regards to subjectivity?}
    \item [Q2.] \textit{How do annotators of diverse backgrounds approach news reports?}
    \item [Q3.] \textit{Should current subjectivity annotation guidelines be further developed to account for morphologically-rich, socially complex, and culturally-specific content?}
\end{enumerate}

To address \textit{Q1}, we confirm that news statements either reflect an accurate description of the entities and events being reported (objective view) or convey the reporter's personal judgments and predictions about the impact of the news (subjective view). The subjective view is typically driven by political, historical, and cultural biases and subjectivities of the reporter or the news agency. We recommend hiring annotators aware of various subjectivity affecting the news being reported to ensure neutral annotations. %Additionally, outlining the construction of the sentences being annotated 

To address \textit{Q2}, we affirm that annotators' political, historical, and cultural backgrounds significantly influence their understanding of the news articles and consequently their judgments. We recommend giving the annotators the option to abstain when they cannot judge sentences. This can be compiled in the annotation tool design by adding the label ``Others'' with the ability to provide justification, forming open-ended annotations that would be more valuable for analysis and validation.

To address \textit{Q3}, we highlight the following points:
\begin{itemize}[noitemsep]

    \item \textit{Semantic Curation for Data}: We endorse the importance of carefully preparing data for annotation for ensuring accurate results. The processing pipeline, including the segmenter, must consider both syntactic and morphological aspects of the sentences. Furthermore, focusing on annotating self-contained and concise sentences will enhance the overall quality of the annotations.

    \item \textit{Abstention with Open-ended Annotations}: We recommend allowing annotators to abstain when they cannot judge sentences. This is mainly because not all sentences must be subjective or objective, some are neutral or ambiguous. This can be implemented in the annotation tool by adding an ``Others'' label with the option to provide justification. Additionally, requesting the rationale behind annotations would enhance their value. Such open-ended annotations would be more valuable for analysis and validation.
    
    \item \textit{Domain-specific Training for Annotation}: We emphasize the importance of specifying the data source in the annotation guidelines. For news reporting, annotators should be trained to distinguish between factual statements and text influenced by biases, as this fine distinction separates objective from subjective sentences.

    \item \textit{Validation Phase for Annotation}: We highlight the importance of implementing a validation phase where annotators can meet and discuss their annotations to minimize discrepancies due to subjectivity by looking at different opinions. 
    % This annotation exercise in the set showed a significant shift from the preliminary annotation, to the final distribution based on majority vote. Objective sentences initially were labeled as such comprising of 767, and decreased to 509 in the final majority vote. The subjective sentences initially labeled as such increased from 77 to 336 in the final majority vote. The points of initial disagreements were further rectified after inter-annotator engagement and discussion, revising the guidelines, and outlining the construction of the sentences being annotated. In the final batch of annotations, annotators experienced less of a discrepancy from the initial annotations, to the final majority vote. We anticipate this to be a result of further refining the statements, and ensuring that the text being annotated is complete, and is not a compilation of sentence fragments, or run-on sentences, that present only one aspect of the entire sentence to be annotated. 

\end{itemize}

\section{Experimental Setup}
\label{sec:experiments}
In this section, we detail the evaluation setup used to benchmark \ds{} and explore the subjectivity of Arabic news articles. 
% We discuss the data setup in Section~\ref{sec:setup_data}, the models used in Section~\ref{sec:setup_models}, and the evaluation measures in Section~\ref{sec:setup_measures}.

%%%%%%%%%%%%%%%%%%%%%%%%%%%%%%%%%%%%%%
\subsection{Data}\label{sec:setup_data}
We used stratified sampling to split the data into training, development, and test sets in a 70:10:20 ratio per class. Table~\ref{tab:data_stats} shows the statistics for each data split.

\noindent\textit{Monolingual Experiments}: We used the training and development splits to fine-tune the pre-trained models. %and tune their hyperparameters. 
The test split was used for evaluation.

\noindent\textit{Multilingual Experiments}: We used three setups for the training data: (i) \textsc{Ar}: \ds{} training set alone, (ii) \textsc{Ml}: the entire multilingual datasets from CheckThat! 2023 \cite{clef-checkthat:2023:task2} and 2024 \cite{clef-checkthat:2024:task2}, and (iii) \textsc{All}: combining both \ds{} training and the entire CheckThat! multilingual datasets. In all setups, we test on \ds{} test set. We show the data statistics in Table~\ref{tab:multi_data_stats}.

\begin{table}[]
    \centering
    \scalebox{0.85}{%
    \begin{tabular}{cl rrr} \toprule
    \textbf{Set} & \textbf{SUBJ} & \textbf{OBJ} & \textbf{All} \\\midrule
    \multirow{2}{*}{\rotatebox[origin=c]{90}{\textsc{Ar}}}
    &Train & 1,055 (66.8\%)  & 1,391	(66.8\%) & 2,446 \\
    &Dev & 201 (12.7\%) & 266 (12.8\%) & 467 \\
    &Test & 323	(20.5\%) & 425 (20.4\%) & 748 \\\midrule

    \multirow{2}{*}{\rotatebox[origin=c]{90}{\textsc{Ml}}}
    &Train	&2,580 (79.1\%)	&4,778 (86.0\%)	&7,358 \\
    &Dev	&357 (11.0\%)	&353 (6.4\%)	&710 \\
    &Test	&323 (9.9\%)	&425 (7.65\%)	&748 \\\midrule
    
    \multirow{2}{*}{\rotatebox[origin=c]{90}{\textsc{All}}}    
    &Train	&3,635 (80.5\%)	&6,169 (85.5\%)	&9,804 \\
    &Dev	&558 (12.4\%)	&619 (8.6\%)	&1,177 \\
    &Test	&323 (7.2\%)	&425 (5.9\%)	&748 \\\bottomrule

    \end{tabular}
    }
    \caption{Statistics of multilingual training data.}
    \label{tab:multi_data_stats}
    \vspace{-0.3cm}
\end{table}

%We used \ds{} with the multilingual datasets from CheckThat! 2023 \cite{clef-checkthat:2023:task2} and 2024 \cite{clef-checkthat:2024:task2}. %, split-wise.

%%%%%%%%%%%%%%%%%%%%%%%%%%%%%%%%%%%%%%
\subsection{Models}\label{sec:setup_models}
We have used three categories of models in our experiments, dummy, pre-trained language models, and large language models. We discuss the models employed under each of these categories in the following:

%%%%%%%%%%%%%%%%%
\paragraph{Simple Models:} To establish reasonably performing baselines, we used three simple models: \textsc{Random}, which assigns labels randomly to sentences; \textsc{Majority}, which assigns the most prevalent label in the dataset to all sentences; and \textsc{SVC}, which employs a C-Support Vector classifier with n-gram features.

%%%%%%%%%%%%%%%%%   
\paragraph{Pre-trained Language Models (PLMs):} We fine-tuned several PLMs to evaluate their performance on the subjectivity task using the transformer toolkit \cite{wolf-etal-2020-transformers}.
%and \hl{tuned different hyper-parameters using the training and development sets}. 

\noindent\textit{Monolingual Experiments}: We fine-tuned \textsc{AraBERT} version 2~\cite{antoun2020arabert} and \textsc{Qarib}~\cite{abdelali2021pretraining}, both of which are initially trained on Arabic datasets.

\noindent\textit{Multilingual Experiments}: We fine-tuned multilingual BERT (\textsc{mBERT})~\cite{devlin-etal-2019-bert} and XLM-RoBERTa base (\textsc{RoBERTa}) %$_{base}$) %and \textsc{RoBERTa}$_{large}$)~
\cite{conneau-etal-2020}.

All these models were fine-tuned using the training dataset of \ds{} or the entire multilingual data from the Subjectivity Task 2 in CheckThat! Lab 2023~\cite{clef-checkthat:2023:task2} and 2024~\cite{clef-checkthat:2024:task2}.

%%%%%%%%%%%%%%%%%
\paragraph{Large Language Models (LLMs):} To align with recent advancements in NLP, we experimented with GPT-4 (version 0314) \cite{achiam2023gpt} in both zero-shot and few-shot setups. For reproducibility, we set the temperature to zero for zero-shot experiments and designed the prompts using concise instructions similar to those given to human annotators when creating \ds{}. We utilized the LLMeBench framework to run our experiments~\cite{dalvi-etal-2024-llmebench}.

%%%%%%%%%%%%%%%%% Prompts
%\paragraph{Prompts, Zero, Few-shots:}
The use and evaluation of LLMs involve prompting and post-processing of output to extract the expected label. For each GPT-4 experimental setup we iterative tried multiple prompts guided by the same instruction and format as recommended in in OpenAI playground. After having an expected prompt, we run complete evaluation. 

\noindent \textit{\textbf{Zero-Shot.}} 
For the zero-shot experiments, we designed prompts by providing natural language instructions that describe the task and specify the expected label.

\noindent \textit{\textbf{Few-Shots.}} 
For the few-shot example selection, we used the maximal marginal relevance-based (MMR) method to construct example sets that are both relevant and diverse \cite{carbonell1998use}. The MMR method calculates the similarity between a test example and the example pool (e.g., training set) and selects $m$ examples (shots). We applied MMR on top of embeddings generated by multilingual sentence-transformers \cite{reimers2019sentence}. We conducted experiments with 3-shot and 5-shot examples.

%%%%%%%%%%%%%%%%%%%%%%%%%%%%%%%%%%%%%%
\subsection{Evaluation Measures}\label{sec:setup_measures}
We evaluate all models' predictions using classification metrics including weighted Precision, Recall, and F1-score for the ``Subjective'' class.
%, along with macro-averaged scores.
\section{Results and Discussion}
\label{sec:results}
% In this section, we discuss the results of the benchmarking experiments, monolingual in Section~\ref{sec:results_mono} and Multilingual experiments in Section~\ref{sec:results_multi}.

\subsection{Monolingual Results}\label{sec:results_mono}
Table~\ref{tab:mono_results} presents the benchmark results of all models trained on the \ds{} training split. \textsc{Gpt-4}, in both zero- and few-shot learning, markedly surpasses all other models across all measures except Recall. The optimal setup for \textsc{Gpt-4} is the 3-shot setup, showing a reasonable improvement compared to the 0-shot and 5-shot setups. Notably, in terms of \emph{Recall}, \textsc{AraBERT} outperforms \textsc{Gpt-4}, which merits further investigation. This could indicates a weakness in \textsc{Gpt-4} in identifying all ``Subjective'' sentences, despite achieving the highest \emph{Precision} scores by more frequently assigning the ``Subjective'' label to sentences.
% Qarib is not performing well!
Clearly, the PLMs perform better when trained on Arabic data, \textsc{Qarib} and \textsc{AraBERT}, using the $F_1$ score.

\begin{table}[h!]
    \centering
    \scalebox{0.9}{%
    \begin{tabular}{l p{.8cm}p{.8cm}p{.8cm}p{.8cm}}\toprule
    Model	&Acc	&P	&R	&F1  \\\midrule
    \textsc{Random}	&0.568	&0.284	&0.500	&0.362 \\
    \textsc{Majority}	&0.500	&0.499	&0.499	&0.497 \\
    \textsc{SVC}	&0.540	&0.517	&0.515	&0.509 \\\midrule
    \textsc{Qarib}	&0.523	&0.519	&0.523	&0.520 \\
    \textsc{AraBERT}	&0.592	&0.582	&\textbf{0.592}	&0.566 \\\midrule	
    \textsc{mBERT}	&0.563	&0.549	&0.563	&0.546 \\
%    \textsc{RoBERTa}$_{base}$	&	&	&	& \\		
%    \textsc{RoBERTa}$_{large}$	&0.568	&0.323	&0.568	&0.412 \\\midrule			
    \textsc{RoBERTa}	&0.568	&0.323	&0.568	&0.412 \\\midrule
    \textsc{Gpt-4}$_{0-shot}$	&0.768	&0.517	&0.507	&0.529 \\
    \textsc{Gpt-4}$_{3-shot}$	&\textbf{0.795}	&\textbf{0.647}	&0.544	&\textbf{0.800} \\
    \textsc{Gpt-4}$_{5-shot}$	&0.785	&0.636	&0.528	&\textbf{0.800} \\
    \bottomrule
    \end{tabular}
    }
    \caption{Results of mono-lingual models on \ds{} dataset.}
    \label{tab:mono_results}
\end{table}

\subsection{Multilingual Results}\label{sec:results_multi}
Table~\ref{tab:multi_results} shows the benchmark results of all models trained on \ds{} and multilingual data.

The performance difference between \textsc{mBERT} and \textsc{RoBERTa} models is generally marginal across each setup. Both models achieve their best performance when fine-tuned with only Arabic data (\textsc{Ar} setup). \textsc{mBERT} shows superior performance in the \textsc{All} setup, whereas \textsc{RoBERTa} excels in the \textsc{Ml} setup, demonstrating its robustness in the absence of Arabic training data.

\begin{table}[h!]
\centering
\scalebox{0.9}{%
    \begin{tabular}{p{.8cm}p{1.7cm} p{.8cm}p{.8cm}p{.8cm}p{.8cm}}\toprule
    Setup & Model	&Acc	&P	&R	&F1  \\\midrule
    \multirow{2}{*}{\rotatebox[origin=c]{90}{\textsc{Ar}}}
    %\multicolumn{5}{l}{\textbf{Training on \ds{}}} \\
    & \textsc{mBERT}      &0.563	&\textbf{0.549}	&0.563	&\textbf{0.546} \\
    & \textsc{RoBERTa}    &\textbf{0.568}	&0.323	&\textbf{0.568}	&0.412 \\\midrule

    \multirow{2}{*}{\rotatebox[origin=c]{90}{\textsc{Ml}}}
    %\multicolumn{5}{l}{\textbf{Training on CheckThat!}} \\
    & \textsc{mBERT}      &0.525	&0.498	&0.525	&0.495 \\
    & \textsc{RoBERTa}	&0.532	&0.505	&0.532	&0.500 \\\midrule

    \multirow{2}{*}{\rotatebox[origin=c]{90}{\textsc{All}}}
    %\multicolumn{5}{l}{\textbf{Training on \ds{} and CheckThat!}} \\
    & \textsc{mBERT}      &0.554	&0.535	&0.554	&0.528 \\
    & \textsc{RoBERTa}	&0.532	&0.502	&0.532	& 0.494\\             
    \bottomrule
    \end{tabular}
}
\caption{Results of multilingual models on \ds{} dataset. Refer to Section~\ref{sec:setup_data} for training setup, ``Setup'' column.}
\label{tab:multi_results}
\end{table}

\section{Annotations with Rationals} 
\label{sec:data_rationale}
% \firoj{I used the term explanation in the intro and abstract.., rational is also good. Can we claim, we manually verified X explanation and out of which we agree Y\%?}

We utilized GPT-4 to validate and rationalize the human subjectivity annotations. Specifically, for each sentence in \ds, we prompted GPT-4 with the sentence and its label, and asked it, as an expert linguist, to \textit{``Write a simple and short explanation''} for its given annotation. We generated explanation in both Arabic and English languages, which we will release along with \ds{} for the community. Table~\ref{tab:explain_prompt} shows the prompt and example output in both languages.

\section{Instruction Dataset} 
\label{sec:data_instruct} 
To instruct-tune LLMs, it is essential to create an instruction dataset. For this purpose, we used GPT-4o to generate instructions for the development and test sets. The generated instructions from the development were then used to assign instructions randomly to the samples in the training dataset. Let \( D_{\text{dev}} \) be the development set. We denote the set of instructions generated by GPT-4o for \( D_{\text{dev}} \) as \( I \), as shown in Equation \ref{eq:instructions}:
\begin{equation}
I = \{ \text{GPT-4o}(x) \mid x \in D_{\text{dev}} \}
\label{eq:instructions}
\end{equation}

Let \( D_{\text{train}} \) be the training set. Instructions from \( I \) are assigned randomly to each sample in \( D_{\text{train}} \), as represented in Equation \ref{eq:train}:

\begin{equation}
\forall x \in D_{\text{train}}, \; \text{assign} \; I_{\text{rand}}(x) \in I
\label{eq:train}
\end{equation}

where \( I_{\text{rand}}(x) \) denotes an instruction randomly selected from \( I \). This ensures that each training sample is paired with an instruction. Note that we kept the instruction from the test set independent. The purpose of using instruction from the dev set to the training set was to reduce the API cost of GPT-4o. 

To create instructions for the development and test sets, we aimed to generate diverse instructions. In Listing \ref{lst:prompt_gpt4_instruct}, we present the prompt used to create these instructions. For different samples, we asked GPT to create various types of instructions, such as \textit{(i)} simple, \textit{(ii)} straightforward, and \textit{(iii)} detailed. We randomly selected one type from the three and used in the placeholder \textit{random\_ins\_type}. The placeholder \textit{sentence} represents the input sentence. 

\section{Conclusion and Future Work}
\label{sec:conclusion}
% In these three batches of annotations, and overall four phases of corpus development, the research presented various opportunities for further research in the field of SSA. Primarily, the annotation process can be further enhanced to adapt to the complexities of the Arabic language both semantically, and culturally, including instances where culturally specific connotations and statements are accounted for. As this study has shown, inter-annotator disagreement and majority vote are both ripe with opportunities and limitations in understanding subjectivity annotation as a task. 
In this study, we propose a large subjectivity dataset for Arabic, consisting of manually annotated news sentences. We provide a detailed discussion of the data collection and annotation process. For the classification experiments, we conducted extensive experiments with PLMs and LLMs to demonstrate the utility of the dataset and system development. Additionally, we provide rationales for each sentence being classified as subjective or objective. Furthermore, we created an instruction-following dataset, which can be used in LLM-based model development. 

Given the complexity of annotation, future research should include more annotators from diverse backgrounds to further enhance the subjectivity annotation process. Our study is preliminary in nature and serves as an initial step towards understanding news media in terms of subjectivity. However, this study presents important considerations for scholars specifically interested in subjectivity and for the field of NLP in general.
\section{Limitations} \label{sec:limitations}
Subjectivity annotation is a complex task, which has also been noticed in other languages. Even though we provided clear guideline in Arabic to make sure that native speakers fully understand the task, however, it still become a challenge for that. Many mturk annotators did not pass our qualified test. It might be because they are from diverse background, and culture, which might have effected the annotation process.

\section*{Ethics and Broader Impact}
We collected news articles from various sources and selected specific sentences for the annotation. It is important to note that annotations are subjective, inevitably introducing biases into our dataset. However, we provided clear annotation schema and instructions aim to minimize these biases. We urge researchers and users of this dataset to remain careful of its potential limitations when developing models or conducting further research. Models developed using this dataset could be invaluable to fact-checkers, journalists, social media platforms, and policymakers.

\section*{Acknowledgments}
The contributions of F. Alam, M. Hasanain, R. Suwaileh and W, Zaghouani were funded by the NPRP grant 14C-0916-210015, which is provided by the Qatar National Research Fund part of Qatar Research Development and Innovation Council (QRDI).

\bibliography{custom}
\bibliographystyle{acl_natbib}

\appendix

\section{Annotation Guidelines}
\label{apndix:guidelines}
For the annotation we adopted and refined the annotation guidelines discussed in \cite{Francesco21}. 
To begin the annotations, annotators of diverse backgrounds were provided with a specific use-cases for subjective and objective sentences that we present in Tables~\ref{tab:news_analysis}, \ref{tab:subj_usecases} and~\ref{tab:obj_usecases}, respectively.

\subsection{Subjective Use Cases}
%A sentence is considered \emph{subjective} when it is based on or influenced by personal feelings, tastes, or opinions. Otherwise, the sentence is considered objective.
We define \emph{subjective} sentences as expressions of feelings, literary tastes, or personal interpretations of topics and events. Below are a few use cases of subjective sentence with examples in Table~\ref{tab:subj_usecases}:

\begin{itemize}[noitemsep]
    \item Sentences expressing personal opinions about events and topics, or containing rhetorical questions, or containing probabilities and expectations and building conclusions on them, e.g., Sentence \#1.

    \item Sentences containing sarcasm or humor, according to the writer's expression, e.g., Sentence \#2.

    \item Sentences encouraging, supporting, or approving an action , e.g., Sentence \#3.

    \item Sentences containing offensive expressions such as racism, tactlessness, etc., e.g., Sentence \#4.

    \item Sentences containing a rhetorical expression and depiction of people and situations, such as ``exaggeration'', that a writer uses to express his or her personal opinion, e.g., Sentence \#5.
    
\end{itemize}

\begin{table*}[]
    \centering
    \scalebox{0.95}{%
    \begin{tabular}{l p{6.4cm}p{6.6cm}} \toprule
    \#& Sentence & Translation \\\midrule

    1	  
    & \begin{RLtext} \footnotesize
    دمرت السلطات الصينية آلاف المساجد في شينجيانغ، حسبما ذكر مركز أبحاث أسترالي الجمعة، في أحدث تقرير عن انتهاكات واسعة لحقوق الإنسان في الإقليم المضطرب.\end{RLtext}
    %\textcolor{blue}{\begin{RLtext} \footnotesize\end{RLtext}}
    %\begin{RLtext}.\end{RLtext}
    & Chinese authorities have destroyed thousands of mosques in Xinjiang, an Australian think tank said Friday, in the latest report on widespread human rights violations in the volatile region. \\\\ %\textcolor{blue}{volatile region}. 

    2	  
    & \begin{RLtext} \footnotesize
    وأعلن الاحتلال في أوقات المسيرات خلال الأعوام السابقة عن رفع حالة الاستنفار العسكري على الحدود لأكثر من أسبوعين، خشية اختراق الحدود من دول الطوق باتجاه الأراضي الفلسطينية، والذين من ضمنهم مئات من المتضامنين الدوليين.
    \end{RLtext} 
    %& \textcolor{blue}{The occupation} announced during marches throughout recent years in reducing military presence in borders for more than two weeks for fear of \textcolor{blue}{breaching the borders} from the countries of the enclave towards the Palestinian territories, including hundreds of international solidarity activists. \\\\
    & The occupation announced during marches throughout recent years in reducing military presence in borders for more than two weeks for fear of breaching the borders from the countries of the enclave towards the Palestinian territories, including hundreds of international solidarity activists. \\\\

    3	  
    & \begin{RLtext} \footnotesize 
    هالة صدقي تلجأ إلى الديانة الإسلامية لتحقيق رغبتها وهذه التفاصيل - مشاهير عالمية كم تقاضى جونقكوك من فرقة  \LR{BTS} للغناء في إفتتاح مونديال قطر؟ - المشاهير العرب مفاجأة من العيار الثقيل!! 
    \end{RLtext} 
    & Hala Sudqi resorts to the Islamic religion to achieve her desires and these are the details – celebrities globally how much does Jongkok from BTS make for singing in the opening ceremony of FIFA in Qatar? – Arabic celebrities, a big surprise! \\\\

    4	  
    & \begin{RLtext} \footnotesize 
    وتبين أيضا أن زوجة الجاني سبق ان قامت بقطع المساعدة المالية عن ابنهما، وأن هذا الأخير قام بالتآمر على والديه عن طريق حشو المسدس بالرصاص، وهو عالم بما دأب عليه أبوه من عادة تهديد أمه بالقتل عن طريق ذلك المسدس الفارغ، فإن نفذ تهديده مرة واحدة فسيتخلص من أمه وأبيه بضربة، أو رصاصة واحدة. \end{RLtext} 
    & It was also found that the perpetrator's wife had previously cut off financial assistance from their son, and the son had conspired against his parents by loading the gun knowing that his father usually threatened his mother with death using an empty gun. If the father executes his threat one more time, then the son will be rid of both his parents at once or should we say with one bullet. \\\\

    \bottomrule
    \end{tabular}
    }
    \caption{Examples of news sentences. 
    % \rs{these sentences are not from \ds{}}
    }
    \label{tab:news_analysis}
\end{table*}

\begin{table*}[]
    \centering
    \scalebox{0.95}{%
    \begin{tabular}{l p{6.4cm}p{6.6cm}} \toprule
    \#& Sentence & Translation \\\midrule

    1	  
    & \scriptsize \begin{RLtext} 
    والدليل اغتيال البروفيسور التونسي محمود عبد القادر البزرتي وكان قد فك الشيفرة الوراثية للفيروس وعزلها مخبريا واعادة تشكيلها ليتحول الفيروس الى لقاح مضاد وذلك أغضب أمريكا فقامت باغتياله إنها باختصار جريمة بحق البشرية. \end{RLtext}
    & \footnotesize The evidence is the assassination of Tunisian professor Mahmoud Abdel Qader Al-Bazrti, who had decoded the genetic code of the virus, isolated it in a laboratory, and reconstructed it so that the virus could be transformed into an anti-vaccine. This angered America, so it assassinated him. It is, in short, a crime against humanity. \\\\

    2	  
    & \scriptsize \begin{RLtext} لنعود كل سنة مع أول قطرة غيث إلى نقطة الصفر نشكو لهم من انسداد البالوعات فيقولوا لنا عليكم بالسباحة!!!! \end{RLtext} 
    & \footnotesize Let us return back every year with the first drop of rain to ground zero and complain to them about the clogged drains, and they tell us that you should go swimming!!!! \\\\

    3	  
    & \scriptsize \begin{RLtext} هذا يعني حماقة أو بيستحمق الآخرين، فنحن نعتقد من واجبنا شرعاً أن نقاوم هذا الاحتلال بكل ما أوتينا من قوة ونعاقبه بنفس الطرق التي هو يستخدمها ضدنا. \end{RLtext} 
    & \footnotesize This means foolishness or fooling others. We believe it is our legal duty to resist this occupation with all our might and punish it with the same methods it uses against us. \\\\

    4	  
    & \scriptsize \begin{RLtext} كما سقط مبارك وسيسقط السفيه السيسي، وخطيئة أى ديكتاتور أن ينظر إلى الشعب من خلال عصابته المنتفعة، وأن يصدق نفاقهم ويكذب علامات الغضب الثوري. \end{RLtext} 
    & \footnotesize Mubarak also fell and the foolish Sisi will fall, and it is the sin of any dictator to look at the people through his beneficial gang, believe their hypocrisy and deny the signs of revolutionary anger. \\\\

    5	  
    & \scriptsize \begin{RLtext} لا تظلم أحداً، فالظلم نار لا تنطفئ في قلب صاحبها، ولو مرّت عليه الأعوام. \end{RLtext} 
    & \footnotesize Do not oppress anyone, for injustice is a fire that will never be extinguished in the heart of its perpetrator, even if years have passed. \\\\

    \bottomrule
    \end{tabular}
    }
    \caption{Use cases of ``subjective'' sentences. }
    \label{tab:subj_usecases}
\end{table*}

\subsection{Objective Use Cases}
We define \emph{objective} sentences as a presentation of facts, events, and topics based on factual data. Below are a few use cases of objective sentence with examples in Table~\ref{tab:obj_usecases}:

\begin{itemize}[noitemsep]
    \item Sentences containing news (Sentence \#1), facts (Sentence \#2) and laws (Sentence \#3) conveyed by the writer of the sentence.

    \item Sentences describing the writer's feelings or emotions without expressing any opinions on any topic, e.g., Sentence \#4.

    \item Sentences containing opinions, claims, feelings, or viewpoints attributed to a third party other than the writer, e.g., Sentence \#5.

    \item Sentences conveying the writer's comments without explicitly stating any personal conclusion, interpretation, or expression of a personal opinion, so that the discussion is left open, e.g., Sentence \#6.

    \item Sentences stating conclusions reached by the writer of the sentence, without expressing his personal position or opinion, or they are justified by hypotheses that are not related to personal opinions, e.g., Sentence \#7.

    \item Sentences referring to an individual by a well-known nickname that was not given by the writer, e.g., Sentence \#8.

    \item Common expressions and examples or sayings, e.g., Sentence \#8.
    
\end{itemize}

\begin{table*}[]
    \footnotesize
    \centering
    \scalebox{0.95}{%
    \begin{tabular}{l p{6.4cm}p{6.6cm}} \toprule
    \#& Sentence & Translation \\\midrule

    1	  
    & \scriptsize \begin{RLtext}
    و أضافت نفس المصادر أن هناك أحزاب رفضت هذا الاقتراح نظرا لأن بن عيسى غير ملم بالوضع الاقتصادي الحرج التي تمر به تونس. \end{RLtext}
    & \footnotesize The same sources added that there are parties that rejected this proposal because Ben Aissa is not familiar with the critical economic situation that Tunisia is going through.  \\\\

    2	  
    & \scriptsize \begin{RLtext} ويمثل الفيروس خطرا بشكل خاص على كبار السن وعلى مَن يعانون من مشاكل صحية، ولدى إيطاليا واحدة من أكبر نسب كبار السن في العالم. \end{RLtext} 
    & \footnotesize The virus represents a particular danger to the elderly and those suffering from health problems, and Italy has one of the largest proportions of elderly people in the world. \\\\

    3	  
    & \scriptsize \begin{RLtext} في حالة وجود منظمة غير حكومية مرخص لها بتصدير أو إعادة تصدير الخدمات لسوريا بموجب هذا القسم في التاريخ السابق لتاريخ سن هذا القانون، فإن هذا القسم ينطبق على هذه المنظمة في تاريخ سن القانون وبعده إلى الحد نفسه وبنفس الطريقة التي كان ينطبق بها هذا القسم على هذه المنظمة في التاريخ السابق لتاريخ سن هذا القانون. \end{RLtext} 
    & \footnotesize If a non-governmental organization licensed to export or re-export services to Syria under this section existed on the date prior to the date of enactment of this Act, this section shall apply to such organization on and after the date of enactment of this Act to the same extent and in the same manner as this section applied to such The organization on the date prior to the date of enactment of this law. \\\\

    4	  
    & \scriptsize \begin{RLtext} وكلما عدت من التحقيق والمحاكم منهكة إلى زنزانتي أجدد عهدي لله بأنني رغم كل التضييقات سأسير حتى مماتي على نفس الدرب لأنني نذرت نفسي للمسجد الأقصى. \end{RLtext} 
    & \footnotesize Whenever I return from the investigation and the courts exhausted to my cell, I renew my pledge to God that despite all the restrictions, I will walk the same path until my death because I have vowed myself to Al-Aqsa Mosque. \\\\

    5	  
    & \scriptsize \begin{RLtext} وتقول إنه في المجتمع العلمي، هناك تعاريف متضاربة حول موعد تحقيق مناعة القطيع. \end{RLtext} 
    & \footnotesize She says that in the scientific community, there are conflicting definitions about when herd immunity will be achieved. \\\\

    6	  
    & \scriptsize \begin{RLtext} لقد كانت جميلة اسما على مسمى، انطلقت بأحلام الزهور فتعلمت الخياطة والرقص وكانت تحلم بأن تكون مصممة أزياء، ولكن القدر قادها لتكون مصممة أجمل نضال في تاريخ البشرية. \end{RLtext} 
    & \footnotesize She was beautiful by her name. She started out with dreams of flowers, learned sewing and dancing, and dreamed of being a fashion designer, but fate led her to become the designer of the most beautiful struggle in human history. \\\\

    7	  
    & \scriptsize \begin{RLtext} في حال تعرض الفيروس لدرجة حرارة ٢٦ أو ٢٧ سوف يُقتل لذا لا يعيش في المناطق الحارة.\end{RLtext} 
    & \footnotesize If the virus is exposed to a temperature of 26 or 27, it will be killed, so it does not live in hot areas. \\\\

    8	  
    & \scriptsize \begin{RLtext} وجاء ليصب في مصلحة القوى اليمينية الرافضة لمفهوم الوحدة، مثل تيار لوبان في فرنسا، أو حزب الاستقلال البريطاني، غير أن رافضي الدستور الفرنسيين لم يكونوا جميعاً من التيار اليميني بل كانوا أيضاً من التيار اليساري، وسبب رفضهم يرجع إلى انتقادهم للسياسة الاقتصادية المقترحة التي لا تهتم بتأمين ضمانات اجتماعية كافية. \end{RLtext} 
    & \footnotesize It came to serve the interest of the right-wing forces that reject the concept of unity, such as the Le Pen movement in France, or the British Independence Party. However, the French opponents of the constitution were not all from the right-wing movement, but rather they were also from the left-wing movement, and the reason for their rejection is due to their criticism of the proposed economic policy that does not care about Providing adequate social guarantees. \\\\

    9	  
    & \scriptsize \begin{RLtext} الأدعية المأثورة: ``اللهم إني أعوذ بك من البرص والجنون والجذام وسيئ الأسقام وأعوذ بك من همزات الشياطين وأعوذ بك رب أن يحضرون وصلِّ اللهم على سيدنا محمد وآل محمد'' \end{RLtext} 
    & \footnotesize Supplications: ``O God, I seek refuge in You from leprosy, madness, leprosy, and bad diseases, and I seek refuge in You from the instigations of the devils, and I seek refuge in You, my Lord, from their presence, and may God bless our master Muhammad and his family.'' \\\\

    \bottomrule
    \end{tabular}
    }
    \caption{Use cases of ``objective'' sentences. }
    \label{tab:obj_usecases}
\end{table*}

According to the guidelines, subjectivity is defined as follows: ``A sentence is considered subjective when it is based on or influenced by personal feelings, tastes, or opinions. Otherwise, the sentence is considered objective''~\cite{Francesco21}.
% (QCRI, 2023). 
To further illustrate subjective and objective labeling, the guideline presented examples as shown in Table~\ref{tab:guideline_examples}.

\begin{table}[]
\centering
% \resizebox{\columnwidth}{!}{
\scalebox{0.9}{%
\begin{tabular}{@{}lp{6cm}l@{}}
\toprule
\multicolumn{1}{c}{\#} & \multicolumn{1}{c}{\textbf{Sentence}} & \multicolumn{1}{c}{\textbf{Label}} \\ \midrule
1 & \begin{tabular}[c]{@{}p{6cm}@{}}India, who was the bridesmaid at the King’s wedding to Princess Diana in 1981, could not be seen in the footage, but filmed the video as she walked through the royal residence grounds.\end{tabular} & OBJ \\
2 & \begin{tabular}[c]{@{}p{6cm}@{}}It is a sad truth that many of the villages in this region of Portugal are dying.\end{tabular} & SUBJ \\ \bottomrule
\end{tabular}
}
\caption{Examples of subjective and objective sentences. %presented in the guideline
}
\label{tab:guideline_examples}
\end{table}

In the first case, the sentence offers a historical event that is devoid of opinions and personal commentaries. However, the second case designate the ``villages in this region of Portugal'' as ``dying'', which is further qualified as a ``sad truth.'' In light of these examples, the guidelines further present five cases of subjectivity, and seven cases of objectivity as reported in Table~\ref{tab:examples_of_subj_obj}.

\begin{table*}[]
\centering

% \resizebox{\textwidth}{!}{
\scalebox{0.85}{%
\begin{tabular}{@{}lp{14cm}l@{}}
\toprule
\multicolumn{1}{c}{\textbf{C\#}} & \multicolumn{1}{c}{\textbf{Description}} & \multicolumn{1}{c}{\textbf{Label}} \\ \midrule
1 & \begin{tabular}[c]{@{}p{14cm}@{}}Explicit reporting of author’s personal opinion, use of rhetorical questions, speculations, etc.\end{tabular} & SUBJ \\
2 & \begin{tabular}[c]{@{}p{14cm}@{}}Use of sarcasm and irony in expressions attributed to the author of the sentence.\end{tabular} & SUBJ \\
3 & \begin{tabular}[c]{@{}p{14cm}@{}}Evidence of exhortations of personal auspices made by the author of the sentence\end{tabular} & SUBJ \\
4 & \begin{tabular}[c]{@{}p{14cm}@{}}Use of discriminatory and/or downgrading expressions and language. \end{tabular}& SUBJ \\
5 & \begin{tabular}[c]{@{}p{14cm}@{}}Use of rhetorical figures, hyperboles, explicitly used by the author to convey their opinion.\end{tabular} & SUBJ \\
6 & Reporting on historical facts or news quoted by the author. & OBJ \\
7 & \begin{tabular}[c]{@{}p{14cm}@{}}Descriptions of the personal feelings, emotions, or moods of the writer without conveying opinions on other matters.\end{tabular} & OBJ \\
8 & \begin{tabular}[c]{@{}p{14cm}@{}}Expressions of opinions, claims, emotions, or point of views explicitly attributable to a third-party. This includes quotes within sentences, whether at the beginning, end, or both.\end{tabular} & OBJ \\
9 & \begin{tabular}[c]{@{}p{14cm}@{}}Comments made by the author of the sentence that do not draw any conclusion, leaving personal interpretations and opinions out of the sentence construction.\end{tabular} & OBJ \\
10 & \begin{tabular}[c]{@{}p{14cm}@{}}Factual conclusions made by the author that neither convey  any stance or personal opinion nor justified by a personal hypothesis.\end{tabular} & OBJ \\
11 & References to individuals with a well-known nickname or title. & OBJ \\
12 & Common expressions or proverbs. & OBJ \\ \bottomrule
\end{tabular}
}
\caption{Different cases of subjectivity and objectivity.}
\label{tab:examples_of_subj_obj}
\end{table*}

In light of these case studies, annotators were required to select ``Yes'', ``No'', or ``Not Sure'' while annotating each task. ``Yes'' indicated that the sentence is subjective. ``No'' indicated that the sentence is objective. ``Not Sure'' indicated a level of uncertainty outlined by one of these two factors: {\em(1)} the sentence has mixed clauses, both subjective and objective statements, with relatively equal weight in their construction; {\em(2)} the sentence is incomplete and cannot be defined as either subjective and/or objective. 

Three annotators participated in this project. The annotators selected for the subjectivity annotation task come from three different Arabic-speaking backgrounds: Egyptian, Yemeni, and Bahraini. Each nationality has its own dialect, its own forms of expression, despite all three being MSA proficient. 
% One of the annotators is equipped with linguistics and computational experiences, while the other two annotators are in the humanities—digital and political.
The educational background of one annotator is in linguistics and computational studies, while the other two have backgrounds in digital humanities and political science.
%The annotators were further given 
We used an in-house deployed online annotation platform to facilitate the annotators in their annotation tasks. The platform is equipped with a tutorial that includes examples.
% contribution access to the QCRI platforms, to further begin the task labelling, conclude the annotations, and reflect on the results of the annotations. 

%%%%%%%%%%%%%%%%%%%%%%%%%%%%%%%%%%%%%%%%%%
\section{Annotation Challenges}
%Wajdi
Annotating for subjectivity presents significant challenges, especially when conducted via crowdsourcing platforms. One major obstacle is the lack of shared cultural, linguistic, and experiential backgrounds among annotators. In this study, annotators hailed from diverse Arabic-speaking regions like Egypt, Yemen, and Bahrain, each with its own dialect and modes of expression. Their educational backgrounds also varied, ranging from linguistics and computational studies to digital humanities and political science. Such disparities inevitably influenced how annotators interpreted sentences and judged subjectivity, leading to disagreements. While the annotation guidelines were adapted to Arabic, individual linguistic nuances and lack of a common native language among annotators still posed challenges. Furthermore, subjective annotations are inherently influenced by individual biases, standpoints, and opinions, which are difficult to control in a crowdsourced setting. Achieving reliable annotations required iterative refinement of guidelines, pilot studies, qualifications tests, and ongoing quality checks – underscoring the complexities of crowdsourcing subjective annotations across diverse annotator pools.
This study is comprehensive in its approach towards news material, and experienced series of improvements across the four phases of corpora generation and subjectivity annotations. However, there are specific challenges that emerged while annotating.
%Wajdi

First, the guidelines presented to the annotators are in English, representing general descriptions of subjectivity and objectivity, often universally agreed upon between Arabic and English speakers. However, the absence of guidelines in Arabic offers limitations to annotators. 
In the following examples, we highlight a few challenging cases. 
%Consider the below examples:
\begin{enumerate}
\itemsep0em 
    \item \begin{RLtext}
\footnotesizeنسب مشاهدة أعلى من نسب
        متابعةخطاب سعيّد\end{RLtext}
    \textbf{Translation:} The percentage of viewing is higher than those follow Saeed’s speech.	\\
    \textbf{Label:} OBJ	\\ %Majority Annotation: ``No''
    In this %the first 
    sentence, annotators all agreed that the sentence is objective, following case \#6 listed in Table 1. The statement as it stands provides a reporting of statistics, and does not account for any personal expressions on whether the viewing of the first clip was due to a personal standpoint as opposed to viewing of the second clip.
    \item \begin{RLtext}\footnotesize حسبي الله ونعم الوكيل شي محزن
        عندما تصبح مشاعر الناس للضحك والوناسه
        بلا احساس ولا إنسانية.\end{RLtext}
    \textbf{Translation:} Allah suffices me, for He is the best disposer of affairs… It is sad when people's feeling are mocked. People have become insensitive and inhumane. \\
    \textbf{Label:} SUBJ	\\%Majority Annotation: ``Yes'' \\
    %The second 
    This sentence resulted in overwhelming agreement by all three annotators meriting the label of ``subjective''. Here, subjectivity was a result of the second and third clause of the sentence. The first clause in this sentence is a common phrase used by Muslims and if seen independent of the rest of the sentence, would therefore be labeled as objective following case \# 12 as see in Table \ref{tab:examples_of_subj_obj}. However, the remaining two clauses indicate a personal opinion of people engaging in fun and mockery are further being inhumane and insensitive. This in light of the sentence, renders the entire expression as subjective, following cases \#1, 4, 5, and 7. 
    \item \begin{RLtext}\footnotesize	اعتبر أن حكم محكمة العدل ليس له أي تأثير على موضوع النزاع.\end{RLtext}
    \textbf{Translation:} I consider that the ruling of the Court of Justice has no bearing on the subject matter of the dispute.	\\
    \textbf{Label:} SUBJ	\\%Majority Annotation: 1 ``Not Sure''; 2 ``Yes'' \\
    %The third 
    This sentence displays unique instances in which annotators disagreed. The point of disagreement will be further discussed in the results section. However, in light of annotation, there are important factors to consider here. The sentence preliminary translated begins with ``I consider'' meaning a person is expressing a view based on a specific result. Another reading of this sentence could be ``Consider that the ruling'' indicating a speaker speaking directly to the reader. Another reading of this sentence, if it was incomplete, would be ``[He] considered''. The absence of diacritics in this case renders subjectivity annotation as a difficult task to accomplish, further displaying how an incomplete sentence, here in terms of diacritics and grammar, cannot be defined straightforwardly as either subjective and/or objective.
    \item \begin{RLtext}\footnotesizeوتقول الصين إن معسكراتها هي 
 مراكز للتدريب المهني
وهي ضرورية لمكافحة الفقر والتطرف، فيما نفى المتحدث
باسم الخارجية وانغ وينبين نفياً قاطعاً وجود معسكرات
معسكرات اعتقال في شينجيانغ وشكك في مصداقية تقرير  معهد السياسة الاستراتيجية. \end{RLtext}    
    \textbf{Translation:} China says its camps are vocational training centers that are necessary to combat poverty and extremism, while foreign ministry spokesman Wang Wenbin categorically denied the existence of "concentration camps" in Xinjiang and questioned the "credibility" of the Strategic Policy Institute's report.	\\
    \textbf{Label} OBJ	\\%Majority Annotation: 1 ``No''; 1 ``Not Sure''; 1 ``Yes'' \\
    This %fourth 
    sentence also presents a point of contention among annotators. Upon an initial reading, one annotator considered this sentence as objective, indicating the quotes as points of references, and the statement as an expression of news. A second annotator, indicated that this sentence is subjective, by claiming the construction of the sentence renders terms such as ``extremism'' or \<\footnotesize``تطرف''> as points of contention and expressions of authorial opinions, in addition to the ``categorically denied''. The Arabic equivalent of ``categorically denied'' holds a much stronger linguistic, political, and social weight indicating that this behavior is an extreme form of exaggeration. The third annotator labeled this statement as ``not sure'' seeing as the terms that have been quoted, within the sentence structure, may indicate a news reference, but could also be used as forms of indirect hyperbole or commentaries on the situation itself. What is deemed as extremism in this sentence, could also be seen as forms of discrimination and xenophobia. \\
    Second, the annotators coming from different backgrounds, having different experiences, standpoints, opinions, also shaped their subjectivity while annotating and labeling the data. We anticipate that this would still happen even if the annotators are sociolinguists. \\
    Third, majority vote and agreement is necessary for a study such as this present work. However, it may also influence the results of the annotations. For example, in this
    %fourth 
    sentence, % presented in this section, 
    users disagreed across the board in their annotation. When an expert annotator was requested to weigh in on this disagreement, the expert deduced that this sentence is objective, following case \#3 of the tutorial and guidelines: ``Case 3: A sentence is objective if it expresses an opinion, claim, emotion or a point of view that is explicitly attributable to a third-party.''    
\end{enumerate}

These challenges may present limitations to the study of subjectivity annotations, however, they also present opportunities for methodological considerations and further research to fill these gaps. Despite these challenges, the study resulted in significant insights into subjectivity annotation, which are further discussed in the next section.

\section{Prompt}
\label{sec:app_prompt}
The performance of the model is highly dependent on the prompting strategy. Designing optimal prompts for each task is challenging and requires multiple iterations. Depending on the prompt, the output varies across all instances of the dataset. For the subjectivity task in this study, we experimented with \textit{(i)} zero-shot and few-shot methods for label generation, \textit{(ii)} generating explanations, and \textit{(iii)} generating the instruction dataset. In Table \ref{tab:explain_prompt}, we provide examples of prompts in Arabic and English for generating explanations, along with the provided sentences and their labels.

\begin{lstlisting}[caption={Prompt to create instructions.}, label=lst:prompt_gpt4_instruct]
prompt = f"You are an expert in creating instruction datasets to train AI models. \
Here, our idea is to create an instruction dataset for a subjectivity detection task. \
The task is to determine whether a sentence is subjective or objective. \
Write a `{random_ins_type}' instruction for this `{sentence}'. Do not include the sentence in the instruction."
\end{lstlisting}

\begin{table*}[t]
    \centering
    \scalebox{0.85}{%
    \begin{tabular}{l p{12.5cm}}\toprule
        Input Prompt & 
        \begin{RLtext} \footnotesize  أنت خبير لغوي. تستطيع فهم التعبير الذاتي في جمل المقالات الإخبارية. اشرح لماذا الجملة التالية: `ومع أن أداء القوات المصرية كان نصرًا لا يمكن لمنصف أن يماري فيه، فإن نظام أنور السادات كان امتدادًا للنظام العربي الرسمي، الذي استولدَه الاستعمار الغربي، وجعله بمثابة غشاوة استقلال خادعة تعمي عيون وبصائر الشعوب العربية، وتطمس على قلوبها فتغفل عن واقعها المرير'. تعتبر `ذاتية'. اكتب شرح موجز وبسيط. \end{RLtext} \\[.2cm]
        GPT Response &
        \begin{RLtext} \footnotesize الجملة المذكورة تعتبر ذاتية لأنها تحتوي على تقييم شخصي لأداء القوات المصرية ونظام أنور السادات. الكاتب يعبر عن رأيه الشخصي بأن أداء القوات المصرية كان نصرًا لا يمكن الجدال فيه، ولكنه ينتقد نظام السادات بوصفه امتدادًا للنظام العربي الذي أنشأه الاستعمار الغربي. الكاتب يستخدم عبارات مثل `غشاوة استقلال خادع' و`تعمي عيون وبصائر الشعوب العربية' للتعبير عن وجهة نظره الشخصية حول النظام السياسي في مصر والعالم العربي. هذه العبارات تظهر تحيز الكاتب وتقييمه الشخصي للأحداث، مما يجعل الجملة ذاتية. \end{RLtext} \\ \midrule

        Input Prompt & 
        
        You are an expert linguist. You can understand the subjectivity of sentences in news articles. Explain why the  sentence is ``Subjective'':
        \begin{RLtext} \footnotesize `ومع أن أداء القوات المصرية كان نصرًا لا يمكن لمنصف أن يماري فيه، فإن نظام أنور السادات كان امتدادًا للنظام العربي الرسمي، الذي استولدَه الاستعمار الغربي، وجعله بمثابة غشاوة استقلال خادعة تعمي عيون وبصائر الشعوب العربية، وتطمس على قلوبها فتغفل عن واقعها المرير'. \end{RLtext}  
        \LR{Write a simple and short explanation.} \\ \\

        GPT Response & 
        The sentence in question contains subjective language because it includes value judgments and opinions rather than objective facts. Phrases like ``\<\footnotesize نصرًا لا يمكن لمنصف أن يماري فيه>'' (a victory that no fair person can dispute) and ``\<\footnotesize غشاوة استقلال خادعة>'' (a deceptive veil of independence) reflect the author's personal perspective and interpretation of events. Additionally, the use of emotionally charged words such as ``\<\footnotesize تعمي>'' (blind) and ``\<\footnotesize تطمس>'' (obscure) further indicates a subjective viewpoint. The sentence presents the author's opinion on the performance of the Egyptian forces and the nature of Anwar Sadat's regime in relation to Western colonialism and its impact on Arab societies.
        \\\bottomrule
        
    \end{tabular}
    }
    \caption{Prompt for generating explanations for human-annotated sentences}
    \label{tab:explain_prompt}
\end{table*}

\end{document}